\documentclass[a4paper]{article}
\usepackage[english]{babel}
\usepackage{graphicx}
\usepackage{fancyhdr}
\usepackage{setspace}
\usepackage{lineno}	 
\usepackage{lastpage}
\usepackage[numbers,comma,sort&compress,super]{natbib}
\usepackage{nameref}
\usepackage[flushleft]{threeparttable}

\usepackage[utf8]{inputenc} 
\usepackage[T1]{fontenc}
\usepackage{booktabs} 
\usepackage[usenames,dvipsnames]{xcolor}
\usepackage{subcaption}
\usepackage{amsmath}
\usepackage[htt]{hyphenat} 
\usepackage{pgfplots}
\pgfplotsset{compat=1.16}
\tikzset{every picture/.style={/utils/exec={\sffamily}}}
\usepackage{placeins}
\usepackage{ulem}
\newcommand{\tb}[1]{\textcolor{blue}{#1}}

\newcommand{\mytitle}{Towards automatic generation of control structures for Process Flow Diagrams~(PFDs) with Large Language Models}

\newcommand{\myauthor}{Edwin Hirtreiter$^{1}$, Lukas Schulze Balhorn$^{1}$, Artur M. Schweidtmann$^{1,*}$} 
\newcommand{\myauthorshort}{A. M. Schweidtmann}
\author{\myauthor}

\usepackage[colorlinks,linkcolor=blue,citecolor=blue,urlcolor=blue,pdftitle={\mytitle},pdfauthor={Artur M. Schweidtmann}]{hyperref} 
\usepackage{cleveref}

\renewcommand\subsectionmark[1]{}
{
	\fancyhead[L]{\leftmark}
	\fancyfoot[R]{Page \thepage\ of \pageref*{LastPage}}
}

\fancypagestyle{firststyle}
{
	\fancyhead[L]{\copyright \small{\textit{\myauthorshort}}}
	\fancyhead[R]{Page \thepage\ of \pageref*{LastPage}}
	\fancyfoot[L]{\footnotesize \rule{0.25\textwidth}{0.4pt} \newline Corresponding author: $^*$A. M. Schweidtmann, E-mail: a.schweidtmann@tudelft.nl}
	\fancyfoot[C]{}
	\fancyfoot[R]{}
}

\renewenvironment{abstract}{\noindent\textbf{Abstract:}}{}

\newenvironment{acknowledgements}{\noindent\footnotesize\textbf{Acknowledgements}}{}

\usepackage{doi}

\begin{document}
	
	\thispagestyle{firststyle}
	\begin{flushleft}\begin{large}\textbf{Towards automatic generation of control structures for Process Flow Diagrams with Large Language Models}\end{large} \end{flushleft}
	  \myauthor 
	
	\begin{flushleft}\begin{small}
			
			$^1$ Delft University of Technology, \\
			Department of Chemical Engineering, \\
			Van der Maasweg 9, \\
			Delft 2629 HZ, \\
			The Netherlands\\[0.25cm]

		\end{small}
	\end{flushleft}
	
	\begin{abstract}
            Developing Piping and Instrumentation Diagrams (P\&IDs) is a crucial step during process development. We propose a data-driven method for the prediction of control structures. Our methodology is inspired by end-to-end transformer-based human language translation models. We cast the control structure prediction as a translation task where Process Flow Diagrams (PFDs) without control structures are translated to PFDs with control structures. We represent the topology of PFDs as strings using the SFILES 2.0 notation. We pre-train our model using generated PFDs to learn the grammatical structure. Thereafter, the model is fine-tuned leveraging transfer learning on real PFDs. The model achieved a top-5 accuracy of 74.8\% on 10,000 generated PFDs and 89.2\% on 100,000 generated PFDs. These promising results show great potential for AI-assisted process engineering. The tests on a dataset of 312 real PFDs indicate the need for a larger PFD dataset for industry applications and hybrid artificial intelligence solutions.
	\end{abstract}


        \begin{figure}[p!]
        	\centering
            \includegraphics[page=4, scale=0.9]{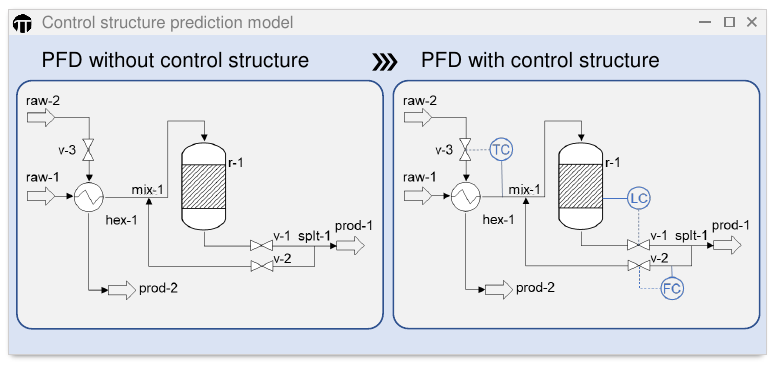}
        	\caption{Graphical abstract}
        \end{figure}

	\newpage
	
\section{Introduction}

Piping and Instrumentation Diagrams~(P\&IDs) are important engineering documents of chemical plants depicting the arrangement of process equipment, valves, piping, control structure, and instrumentation~\cite{Towler.2008}.
In contrast, Process Flow Diagrams~(PFDs) focus on major equipment parts and material streams.
While PFDs are typically used during the early-stage conceptual design phase, P\&IDs are developed in the basic design and detailed engineering phases.
They are essentially the central document in every industrial chemical plant for storing, revising, and exchanging information~\cite{Toghraei.2018}.
The applications of P\&IDs range from engineering and design, to hazard and operability studies~(HAZOPs), construction, operation, maintenance, and decommission~\cite{Toghraei.2018}.

The development of P\&IDs from PFDs is a tedious and time-consuming task that offers great potential to reduce costs and speed-up the development process~\cite{Uzuner.2012}. 
Commonly, process engineers manually develop P\&IDs adopting and modifying schemes from prior projects, design rules, and their experience utilizing Computer-Aided Design~(CAD) software.
However, this traditional development can be laborious because finding, manually adjusting, and transferring suitable technical solutions from old projects can be tedious and error-prone.
Time constraints can lead to the adoption of non-optimal solutions from previous projects and possible alternatives not being considered~\cite{Uzuner.2012}.
Unleashing the potential of computer algorithms assisting engineers in process engineering may help to reduce development times, reduce costs, increase safety, and avoid errors.

Researchers have been working on the automation of process development since the 90s. 
To assist the engineering process during the creation of P\&IDs, multiple rule-based systems have been developed~\cite{Uzuner.2012, Blitz.1994, Obst.2013}.
Modularization approaches of chemical plants commonly provide the underlying framework of rule-based systems and aim to accelerate process development~\cite{FleischerTrebes.2017, Hohmann.2017, Eilermann.2018}.
The method proposed by Blitz~et~al.~\cite{Blitz.1994} asks a user to define certain inputs, such as material properties and process-specific requirements.
Then, a P\&ID is generated based on the user input and the underlying knowledge-based approach, which is implemented as a decision tree. 
Similarly, Uzuner~et~al.~\cite{Uzuner.2012} and Obst~et~al.~\cite{Obst.2013} also utilize a knowledge-based method, which is represented as a hierarchical decision tree. 
Uzuner~et~al.~\cite{Uzuner.2012} first divide the chemical process into modules to reduce the complexity of the design problem.
Secondly, design questions and options are used to guide the user to obtain a P\&ID of the desired module.
While the previous works demonstrate the potential of computer-assisted P\&ID development, they have not yet been broadly adopted by industry. 

One step towards the development of P\&IDs is the synthesis of an appropriate control structure.
In decentralized control, a controller adjusts manipulated variables based on observations of measured variables in order to follow a set point of a controlled variable or optimize another operating objective~\cite{Morari.1980}.
Typically, the development of a plant-wide control scheme starts by analysis of the degrees of freedom (i.e., the number of controllable variables)~\cite{Luyben.1997, Ng.1998}. 
In parallel, control tasks are defined that relate to decentralized operational targets, such as set-point tracking or disturbance rejection of product qualities or flow rates, and overarching economic and ecological objectives, such as maximum product output or minimum energy consumption~\cite{Morari.1980, Luyben.1997, Ng.1998}.
To assist process engineers in developing (decentralized) control structures that specify all controllable variables and achieve the process objectives, several established methods exist including dynamic simulations~\cite{Seborg.2011} and relative gain array methods~\cite{VANDEWAL.2001}. 
In addition, heuristic design procedures~\cite{Luyben.1997} and knowledge-based expert systems~\cite{Williamson.1989, SongPark.1990} have been proposed in the 90s.
However, many expert systems in chemical engineering have not led to the expected major advances~\cite{Venkatasubramanian.2019}.
In particular, rule-based systems are often difficult to set up, maintain, and extend~\cite{Uzuner.2012, Venkatasubramanian.2019}.

Recent research and development in deep learning-based Artificial Intelligence~(AI) applications promise improvements over expert systems revealing an outstanding performance in numerous disciplines, as highlighted by the following examples.
In particular, Natural Language Processing~(NLP), a subfield of AI focusing on natural language, with its powerful models (e.g. GPT-3~\cite{Brown.2020}, T5~\cite{Raffel.2019}) showed breakthrough performance in many natural language tasks outperforming systems that previously used handcrafted rules~\cite{Brown.2020, Raffel.2019, Vaswani.2017, Popel.2020}.
We already see great speedups of AI-assisted workflows in many domains (e.g., deepL for translations or GitHub Copilot software development). Similarly, there are a large number of domains that explore the potential of ChatGPT, although there are no guarantees of correct predictions.
Also, deep learning has outperformed rule-based approaches in organic chemistry.
For example, transformer-based language models can accurately predict reaction outcomes based on string representations of reactions using the Simplified Molecular-Input Line-Entry System~(SMILES) notation~\cite{Weininger.1988, Weininger.1989, Schwaller.2018, Schwaller.2019, Schwaller.2020}.

In the context of process engineering, there exist a few very recent and promising methods, which learn patterns from existing PFDs and P\&IDs~\cite{Zhang.2019, Zheng.2022, Oeing.2022, Vogel.2022}.
Zhang~et~al.~\cite{Zhang.2019} and Zheng~et~al.~\cite{Zheng.2022} use the Simplified Flowsheet-Input Line-Entry System~(SFILES)~\cite{dAnterroches.2006} notation to describe the flowsheet topologies as strings in conjunction with sequence alignment algorithms to identify design heuristics in process diagrams.
Oeing~et~al.~\cite{Oeing.2022} propose an AI-assisted method to predict the subsequent equipment using a Recurrent Neural Network~(RNN).
Similarly, we proposed a methodology for auto-completion of flowsheets based on transformer language models~\cite{Vogel.2022}.  
To enable the use of NLP models, we utilized the SFILES~2.0~\cite{Vogel.2022b} notation.
While previous methods focus on the completion of incomplete process diagrams, there is no method available that enables the generation of PFDs with decentralized control structures directly from basic PFDs without control structures. 

We propose a novel methodology to generate PFDs with decentralized control structures from PFDs without control structures as a first step toward the automatic generation of P\&IDs.
Notably, we provide a conceptual contribution as well as a proof-of-concept which demonstrates that predicting decentralized control structures on synthetic data is feasible.
The underlying idea of our approach is to cast the control structure prediction as a translation task, the source language being the PFD without control structure and the target being the PFD with control structure. 
To leverage the potential of state-of-the-art sequence-to-sequence translation models, based on the transformer architecture~\cite{Vaswani.2017}, we utilize the text-based SFILES~2.0 notation~\cite{Vogel.2022b} to represent the topological information of PFDs.

The remainder of this paper is structured as follows: 
Section~\nameref{sec:Background} describes the fundamentals of the applied natural language model and summarizes the concept of the SFILES~2.0 string representation of chemical processes typologies. 
Thereafter, in Section~\nameref{sec:Data} we describe the data acquisition.
In Section~\nameref{sec:Method} we introduce the transformer model adapted for predicting the control structure.
Afterward, the results are discussed in Section~\nameref{sec:Results} and demonstrated with an illustrative example in Section~\nameref{sec:Examples}.

\section{Background}
\label{sec:Background}

This section summarizes the fundamentals of sequence-to-sequence models for the translation of natural language~(Section~\nameref{subsec:Seq2Seq}). 
In Section~\nameref{subsec:Transformer}, we highlight the transformer architecture as the state-of-the-art deep learning architecture for translation. 
Thereafter, the concept of the SFILES~2.0 notation, which enables a text-based representation of PFDs including control structures, is outlined in Section~\nameref{subsubsec:SFILES2}.
The underlying idea for using SFILES~2.0 in combination with NLP methods was originally proposed by Vogel~et~al.~\cite{Vogel.2022}. Vogel~et~al.~\cite{Vogel.2022} utilizes a generative transformer model consisting of a decoder-only model structure. The decoder is fed with an incomplete PFD, represented as SFILES~2.0, and generates step-by-step the missing parts of the flowsheet. While the previous work proposed a flowsheet autocompletion methodology~\cite{Vogel.2022}, we developed a concept for the prediction of control structures in PFDs. For this purpose, we develop a Sequence-to-Sequence model with an encoder-decoder structure. I.e. we map the PFD without control structure directly to the PFD with control structure.

\subsection{Sequence-to-sequence models}
\label{subsec:Seq2Seq}

Sequence-to-sequence models are machine learning models that map an input sequence to an output sequence. 
They are utilized in numerous NLP tasks, e.g., in translation~\cite{Stahlberg.2020}, text summarization~\cite{Nallapati.2016}, speech recognition~\cite{Chiu.2017}, and image captioning~\cite{Karpathy.2014}. 

\begin{figure}[p!]
	\centering
	\includegraphics[page=1, scale=0.8]{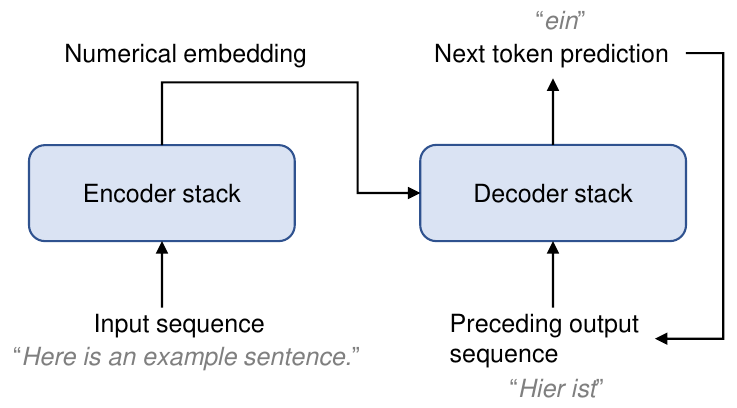}
	\caption{Encoder-Decoder structure of sequence-to-sequence models}
	\label{fig:EncoderDecoder}
\end{figure}

Typically, a sequence-to-sequence model comprises an encoding and decoding stack as depicted in Figure~\ref{fig:EncoderDecoder}. 
During encoding, a numerical embedding of the input sequence is determined, which is subsequently used by the decoder stack to generate the output sequence in an auto-regressive way. 
The decoder iteratively processes the preceding output sequence together with the numerical embedding of the encoder to predict the next token (e.g., a word).
The iterative decoding is stopped as the decoder predicts the end of the sequence as the next token.

During decoding, the decoder stack determines the probabilities for each token in its vocabulary, whereupon the next token of the output sequence is identified utilizing a decoding strategy (e.g., using the greedy or beam search decoding strategy).
Greedy search selects the next token of a sequence based on the highest predicted probability at the decoding step.
The greedy strategy is computationally cheap. 
However, it does not ensure a sequence with a maximal overall probability because sequences with a high probability can also contain some tokens with a low probability.
To mitigate this issue the beam search algorithm was introduced in sequence-to-sequence models (e.g., \cite{Sutskever.2014, Graves.2012, BoulangerLewandowski.2012}). 
Beam search selects and memorizes the $N$-best tokens at every decoding step~creating a tree of possible output sequences.
Every selected token is added separately to the preceding output sequence, and thus the decoder is prompted in total $N$-times to predict the output probabilities of the next tokens. 
Therefrom the decoder selects $N$ tokens with the highest probabilities for the next decoding step~and discards branches with a lower probability. 
In the end either the sequence with the overall highest probability is selected or the $N$-best sequences are returned to a user for selection.
Choosing an appropriate beam size~$N$ is a trade-off between generating sequences with high probabilities and computational cost.

Training of sequence-to-sequence models is typically performed using the cross-entropy loss of the predicted output probabilities of the next tokens and the ground truth~\cite{Raffel.2019}.
With the aid of the computed cross-entropy loss the parameters of the model are adjusted to improve the model performance.
Teacher forcing~\cite{Williams.1989} is commonly applied to correct the model at each decoding step, which forces the model to generate the ground truth corresponding to a given input sequence~\cite{Raffel.2019}.

\subsection{Transformer architecture}
\label{subsec:Transformer}

Originally, the underlying model architectures of sequence-to-sequence models comprised variations of RNNs~\cite{Sutskever.2014}. 
To avoid vanishing or exploding gradients, long short-term memory~\cite{Hochreiter.1997} and gated recurrent neural nets~\cite{Chung.2014} were also introduced. 
Recently, the transformer architecture~\cite{Vaswani.2017}, which is based on the sequence-to-sequence model structure, revolutionized the field of NLP demonstrating breakthrough performances on numerous tasks~\cite{Brown.2020, Raffel.2019, Vaswani.2017}.

The transformer architecture~\cite{Vaswani.2017} is based on the auto-regressive encoder-decoder model structure and was originally proposed to perform translation tasks. 
The transformer model relies entirely on attention mechanisms dispensing any recurrence or convolutions. 
Eliminating recurrence and using attention significantly reduces the number of sequential computations and enables fast parallel processing and model training~\cite{Vaswani.2017}.

Attention, being an important core component of the transformer architecture, enables the model to efficiently capture the meaning of a token depending on the context present in the sequence. 
During model training, the weights of query, key, and value matrices are adjusted to learn the bidirectional context of words in a sequence. 
These matrices are used to compute a query~\textbf{q}, key~\textbf{k}, and value~\textbf{v} vector from the input embedding. 
The resulting vectors are packed to query~$Q$, key~$K$, and value~$V$ matrices to efficiently compute the scaled dot-product attention. 
The implementation of the scaled dot-product attention in the transformer architecture includes a scaling factor~$d_k$ corresponding to the layer size:
\begin{equation}
    \mathrm{Attention}(Q, K, V)=\mathrm{softmax}\left( \frac{QK^T}{\sqrt{d_k}}\right)V.
    \label{eq:Attention}
\end{equation}
To allow the model to learn different representations of a single word, multi-head attention is introduced. 
For this purpose, the queries, keys, and values are linearly projected to different dimensions and processed in parallel by multiple attention heads, which are thereafter concatenated. 
Thus, multi-head attention enables the model model to learn multiple relations of a word within a sentence.

The attention mechanisms are not able to cover any positional information due to the absence of recurrence or convolutions in the transformer architecture. 
Therefore, a positional encoding, utilizing sine and cosine functions, is added to the input and output embedding to provide information about the position in the sequence.

The structure of the original transformer architecture comprises an encoder and a decoder stack each containing six identical layers.
The encoder layers consist of a multi-head attention sub-layer followed by a position-wise fully connected feed-forward network. 
Each sub-layer is succeeded by layer normalization and the addition of a residual connection, which prevent "losing" information from the previous layer and facilitate the gradient flow.
Residual connections are bypass connections around layers, which transfer information from previous layers. 
Without those layers, information from preceding layers may otherwise be diminished by operations in the subsequent layers.
The structure of the decoder stack is similar to the encoder with the difference that the decoder contains two attention sub-layers.
The first attention sub-layer is masked, which limits the decoder to attend only to already generated tokens and prevents the decoder "from glancing into the future".
During training, the model is fed simultaneously with input and target data. Therefore, the masking algorithm prevents the model from simply taking the next token from the target sequence, but learns to predict the next token based on the information in the input sequence and the tokens already generated.
The second attention sub-layer, the encoder-decoder attention layer, performs multi-head attention combining the numerical embedding of the last encoder layer and the results of the preceding self-attention layer.

\subsection{Graph- and text-based representation of process diagrams}
\label{subsubsec:SFILES2}

\begin{figure}[p!]
	\centering
	\includegraphics[page=2, scale=0.8]{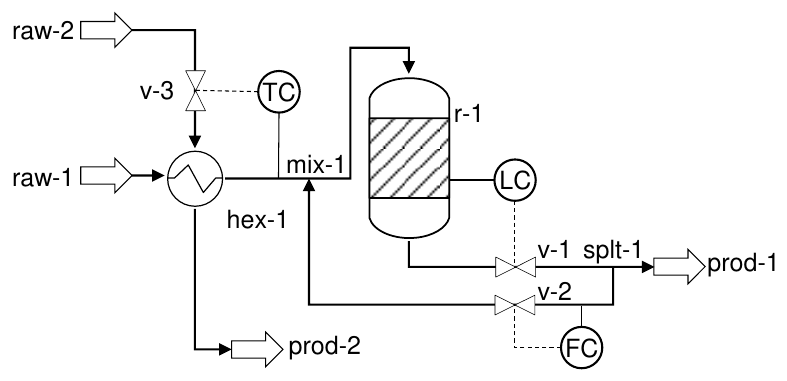}
	\caption{Exemplary chemical process diagram with branching, recycle stream, control units and different mass trains}
	\label{fig:SFILES_Example_Flowsheet}
\end{figure}

\begin{figure}[p!]
	\centering
	\includegraphics[page=3, scale=0.8]{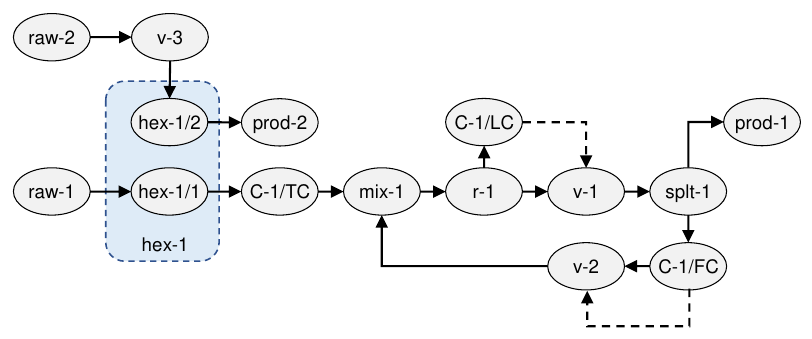}
	\caption{Graph representation of the process diagram in Figure~\ref{fig:SFILES_Example_Flowsheet}}
	\label{fig:SFILES_Example_Graph}
\end{figure}

This section briefly summarizes the graph- and text-based representation of process diagrams as SFILES~2.0~\cite{Vogel.2022, Vogel.2022b}. 
Process diagrams (e.g., PFDs or P\&IDs) of chemical plants can be represented as directed graphs~\cite{Vogel.2022b, Stops.2022}.
Unit operations and control units can be illustrated as nodes in the graph, while material streams and signals are directed edges connecting the nodes.
Figure~\ref{fig:SFILES_Example_Flowsheet} shows an illustrative example process containing a reactor with level control and a recycle loop with flow control. 
This process diagram can be converted to its corresponding graph representation as depicted in Figure~\ref{fig:SFILES_Example_Graph}.
Notably, the two-stream heat exchanger~(hex-1) is split into two nodes to distinguish the two separate material flows, which do not mix inside the heat exchanger.
The control units are stored as nodes like unit operations.

\newpage

SFILES~2.0~\cite{Vogel.2022b} is a text-based representation of process topologies, extending the original SFILES notation as proposed by d'Anterroches~\cite{dAnterroches.2006}.
The SFILES notation is inspired by the SMILES notation, which is used for representing molecules as strings~\cite{Weininger.1988, Weininger.1989}.
With SFILES~2.0 we can efficiently store the topological information of a process graph (e.g., Figure~\ref{fig:SFILES_Example_Graph}) as text, which enables us the application of advanced data processing methods, such as NLP models~(Section~\nameref{subsec:Transformer}).
Converting the graph in Figure~\ref{fig:SFILES_Example_Graph} to the SFILES~2.0 notation with our publicly available \href{https://doi.org/10.5281/zenodo.6901932}{Github repository}~\cite{Vogel.2022c} results in the following string:

\begin{center}
    \texttt{(raw)(hex)\{1\}(C)\{TC\}\_1(mix)<1(r)[(C)\{LC\}\_2](v)<\_2(splt)[(prod)]\\(C)\{FC\}\_3(v)1<\_3n|(raw)(v)<\_1(hex)\{1\}(prod)}.
\end{center}

The SFILES~2.0 notation is read from left to right with two consecutive unit operations respectively control units in parentheses implying a material flow in between.
Branching in the process, for example after the stream splitter~(splt), is represented by putting the individual branches in brackets (here prod), but omitting the brackets for the stream noted last at the branching point (here the recycle flow over C/FC).
Material recycles are included in the SFILES~2.0 notation using a number~\# for the starting point (v) and <\# for the corresponding target (mix).
The heat exchanger is noted twice in the string with a number in braces, indicating that it is the same heat exchanger but two streams enter and leave the equipment.
Independent material streams, such as the utility stream flowing through the heat exchanger compartment hex-1/2, are appended to the SFILES~2.0 string stream separated with n|.
Control units are inserted in the same way as unit operations with subsequent braces indicating the letter code of the instrument.
Signal connections are implemented similarly to material recycles but include an underscore (\_\#, <\_\#).
Currently, the SFILES~2.0 notation is capable of representing flowsheet topologies with their corresponding control structures.
The integration of detailed information on material flows, present components, process dynamics, equipment sizing, special type unit operations, piping information, or operating points is at the moment not available in the SFILES~2.0 notation, but will be subject to further extensions.

\section{Data}
\label{sec:Data}

We use generated data and a dataset of real PFDs with control structures for model training and evaluation.
Section~\nameref{subsec:GeneratedData} describes the generation algorithm of PFDs with decentralized control structures, which are utilized for pre-training the model.
Subsequently, Section~\nameref{subsec:RealData} summarizes the pre-processing of real PFDs with control structures derived from publicly available sources used for model fine-tuning.
It should be highlighted that the generated PFDs and the mined PFDs only contain control structures.
These diagrams are far less detailed than P\&IDs available in the industry containing additional information, such as pipe classes, valves, or instrumentation.

\subsection{Generated data for pre-training}
\label{subsec:GeneratedData}

Typically, NLP models are trained on huge corpora of text that are publicly available on the internet.
For example, \href{https://commoncrawl.org}{Common Crawl} is a publicly available database that extracts around 20~TB of text from the web every month~\cite{Raffel.2019}.
Filtered and cleaned, data from Common Crawl was used as C4~(Colossal Clean Crawled Corpus) dataset with about 750~GB to pre-train the roughly 220~million parameters of the T5-base model~\cite{Raffel.2019}.
Commonly, transfer learning techniques are employed to reduce this massive data demand for new applications \cite{Raffel.2019}. 

Although the SFILES~2.0 notation with its limited, small vocabulary is less complex than natural language, a reasonable amount of pre-training data is necessary to train the randomly initialized weights of the transformer model.
Due to a completely different vocabulary of the SFILES~2.0 compared to natural language, we cannot leverage transfer learning on human language models.
Also, there is no database of PFDs and P\&IDs publicly available.

\begin{figure}[p!]
	\centering
	\includegraphics[page=9, scale=0.85]{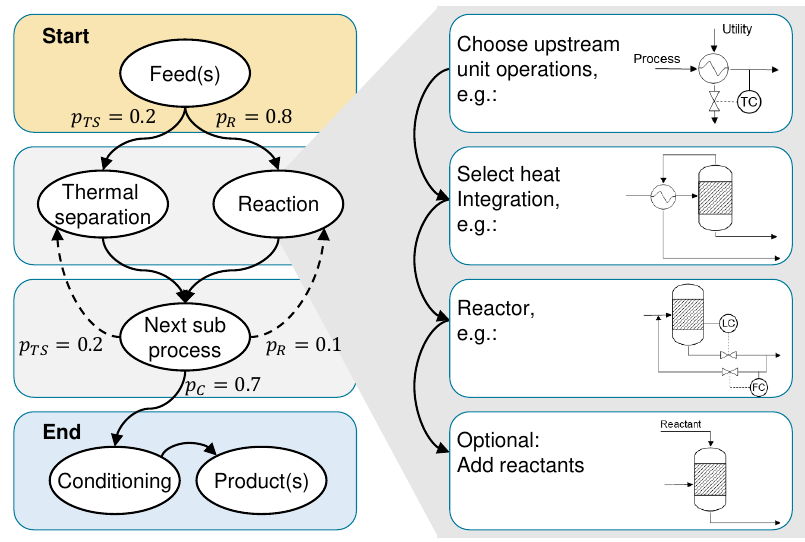}
	\caption{Generation scheme for a PFD with control structure}
	\label{fig:PID_Generation}
\end{figure}

Since data availability is a major limitation of our method, we generate a large set of PFDs with decentralized control structures by extending the previously proposed approach by Vogel~et~al.~\cite{Vogel.2022} for generating several thousand PFDs in a time-efficient way.
This generated dataset is used for pre-training in a transfer learning approach (see Section~\nameref{subsec:ModelTraining} for more details).
The main goal of this data generation step is to obtain a dataset, which is possibly similar to real data. The generated data allows the model to learn the grammatical structure of SFILES~2.0 and to demonstrate the capabilities of the model predicting decentralized control structures.
For this purpose, we implement patterns including decentralized control structures of sub-processes, such as thermal separation or reaction, in which a chemical process is typically divided.
These patterns are thereafter added together to create the PFDs with decentralized control structures of a chemical process consisting of multiple sub-processes. 
The construction of the PFD dataset follows a first-order Markov chain-like sampling process with fixed probabilities.
I.e., the selection of the next sub-process only depends on the current state. 
The probabilities, shown in Figure~\ref{fig:PID_Generation}, for the transition between the sub-processes are selected based on our experience to generate realistic process flowsheets sufficient for pre-training the model.
The utilization of fixed probabilities result in a general structure of the process flowsheet consisting of feed treatment, followed by reaction, thermal separation and final conditioning.
Compared to Vogel~et~al.~\cite{Vogel.2022}, we add control structures based on several basic design heuristics (inspired by~\cite{Towler.2008, Perry.1997, Stichlmair.2020}) for every generated sub-process.
The utilized decentralized control patterns are included in Figure~S1 to S8 in the Supplementary Information.
As illustrated in Figure~\ref{fig:PID_Generation}, we initialize up to three feed streams, which may be pre-processed by inserting heat exchangers, pumps, compressors, or mixing units. 
Thereafter, a Markov transition selects either thermal separation or reaction as the next sub-process.
Exemplary illustrated is the generation pattern for the reaction pattern:
Firstly, upstream unit operations comprising heat exchangers, pumps, and compressors are selected.
Thereafter, present heat exchangers may be pre-selected for heat integration utilizing a reactor outlet stream.
In the next step, one of six stored reactor patterns with an optional material recycle stream is selected.
Optionally, a second or third reactant is fed to the reactor in the final step~completing the reaction pattern.
In general, the patterns have several outlet streams transitioning to the "Next sub-process" state, which lead to multiple Markov transitions to subsequent sub-processes.
Branches are either terminated after reaching the conditioning step~or if the generation algorithm detects a node number exceeding 65, which prevents the generation of very large flowsheet graphs.
Duplicates and process diagrams exceeding a node number of 100 are deleted.
We selected a maximum node number of 100 such that the standard deviation and average node number of the generated dataset are in the same order of magnitude as our real PFD dataset (cf. Table~\ref{tab:Dataset_Properties}).
During the sampling process with fixed probabilities, it can happen that incorrect combinations of decentralized control structures occur, which can ultimately have a negative impact on the accuracy of the trained model.

The resulting process graphs with control structure are automatically converted to SFILES~2.0 using our graph to SFILES~2.0 algorithm~\cite{Vogel.2022c}.
In a subsequent step, the SFILES~2.0 with control structure are converted to SFILES~2.0 without control structure by removing all control instruments (abbrev. C) with their corresponding letter code in braces and signal connections identifiable by an underscore before the number \#.
Finally, the generated pre-training dataset consists of process diagrams without control structure (input data) and process diagrams with control structure (output data).
Table~\ref{tab:Dataset_Properties} summarizes the number of training/validation/test samples for model pre-training and key properties of the dataset.
Besides the number of samples, Table~\ref{tab:Dataset_Properties} shows the average number of nodes~$\overline{n_{nodes}}$, the standard deviation of the number of nodes~$\sigma(n_{nodes})$, and the vocabulary size.
The inclusion of a more diverse set of control patterns and, in particular, letter codes to increase the vocabulary of the generated data set could improve the positive effect of the model pre-training procedure.
The complete generated dataset including the SFILES~2.0 PFDs with decentralized control structures and the corresponding PFDs without control structures are published~\cite{Hirtreiter.2023}.

\subsection{Real data for fine-tuning}
\label{subsec:RealData}

We collected 312 PFD images including control structures from publicly available sources including the google and bing image search engines\footnote{Search keywords: "Piping and Instrumentation Diagram", "Rohrleitungs und Instrumentenfließbild", "P\&ID", "R\&I", "R+I"} and extracted process diagrams from scientific literature using data mining~\cite{Balhorn.2022}.
These process flowsheets originate from various industry and academic domains, such as gas and oil plants, experimental setups, or batch operations.
After the manual selection of process diagrams containing control structure, automatic object detection, and path exploration is performed using our flowsheet digitization algorithm~\cite{Theisen.2022}.
Correcting faulty nodes and edges, adding the letter code of control units, and adding the connectivity of the unit operations and control structures is performed using LabelGraph, which is our custom extension to \href{https://github.com/heartexlabs/labelImg}{LabelImg}~\cite{LabelImg.2018}.
In addition, this manual correction step in LabelGraph ensures the trustworthiness of the mined and digitized process flowsheets. In this step obviously faulty process flowsheets and duplicates are deleted, but due to time limitations, a detailed check could not be performed.
The resulting process graphs are converted to SFILES~2.0 using our code~\cite{Vogel.2022c}.
Then, all control structures are removed to build a dataset consisting of SFILES~2.0 without control structure as our input data and SFILES~2.0 with control structures as our output data.
Key statistics of the dataset are denoted in Table~\ref{tab:Dataset_Properties}. 
The table shows that the standard deviation of the number of nodes in the real data (28) is significantly higher than in the generated data (20) while the average number of nodes is smaller in the real data.
This indicates that the sizes of the process diagrams vary more strongly in the real dataset.
The table also highlights a significantly higher vocabulary size of the real dataset (390) compared to the generated dataset (113).
The reason for this is mainly a diversity of additional, new letter codes, but also other new unit operations, which are not present in the generated data.
To conclude, identifying good measures for the comparison of process diagrams (here, generated data and real data) is difficult. As a first step, we used the average node number, the standard deviation of the number of nodes and the vocabulary size. In the future, an investigation of different measures, such as graph similarity, might be helpful for comparing two different process flowsheets.

\begin{table}[p!]
	\caption{Dataset properties and training~(tr), validation~(val), test~(te) splits used for the experiments}
	\centering
	\begin{tabular}{lcc}
		\toprule
		& Generated data & Real data\\ 
		\midrule
		samples\textsubscript{\textit{tr}} & 100/1000/10000/100000 & 250\\
		samples\textsubscript{\textit{val}} & 1000 & 31\\
		samples\textsubscript{\textit{te}} & 1000 & 31\\
		$\overline{n_{nodes}}$ & 52 & 37\\
		$\sigma(n_{nodes})$ & 20 & 28\\
		vocabulary size & 113 & 390\\
		\bottomrule
	\end{tabular}
	\label{tab:Dataset_Properties}
\end{table}

\subsection{Data augmentation}

Data augmentation methods are commonly applied to datasets to increase their size without the effort of manual labeling and to improve the robustness of machine learning models.
In computer vision, images are rotated, cropped, or distorted to have multiple instances of the original image, which are from the computer's point of view completely different.
In the field of NLP, data augmentation is more difficult since the meaning of the sentence has to be preserved.
Techniques in NLP for data augmentation include e.g., synonym replacement, back-translation, random insertion, deletion, or swapping of words~\cite{Sennrich.2016, Wei.2019, Feng.2021}.

To augment the process diagram datasets, we modify the branching decision in the SFILES~2.0 generation algorithm to create different SFILES~2.0 strings representing the same process diagram~\cite{SchulzeBalhorn.2022}.
This procedure is motivated by significant performance advances when using augmented SMILES in neural networks~\cite{Schwaller.2019, Bjerrum.2017, Tetko.2020}.
When generating augmented (non-canonical) SFILES~2.0, the branching decision is made randomly, whereas in the case of the determination of canonical SFILES~2.0 the branching decision is predetermined by assigning every node of the graph to a unique rank~\cite{Vogel.2022b}.
The resulting augmented SFILES~2.0 is grammatically correct and contains identical information as the canonical SFILES~2.0 and thus describes the same process flowsheet.
During augmentation, only the uniqueness of the SFILES~2.0 representation is lost.
For the augmented model training runs we roughly doubled the training data by generating a second SFILES~2.0 for every PFD in the input dataset.
As an example, the PFD corresponding to Figure~\ref{fig:SFILES_Example_Flowsheet} is represented by the following canonical SFILES~2.0 
\begin{center}
    \texttt{(raw)(hex)\{1\}(mix)<1(r)(v)(splt)[(prod)](v)1n|(raw)(v)(hex)\{1\}(prod)},
\end{center}
which can be augmented to
\begin{center}
    \texttt{(raw)(hex)\{1\}(mix)<1(r)(v)(splt)[(v)1](prod)n|(raw)(v)(hex)\{1\}(prod)}.
\end{center}
To review more examples of augmented SFILES~2.0, we published the augmented dataset together with the non-augmented generated dataset~\cite{Hirtreiter.2023}.

\section{Control structure prediction model}
\label{sec:Method}

In the following section, we provide an overview of the general procedure to predict the control structure of PFDs utilizing a sequence-to-sequence transformer model.
In Section~\nameref{subsec:Tokenization}, we describe the tokenizer that enables the model to process the SFILES~2.0 strings.
Thereafter, key parameters of the utilized transformer architecture are briefly summarized in Section~\nameref{subsec:T5}.

\subsection{Overview}
\label{subsec:Overview}

\begin{figure}
     \centering
     \begin{subfigure}[htb]{\textwidth}
         \centering
         \includegraphics[page=7, scale=0.85]{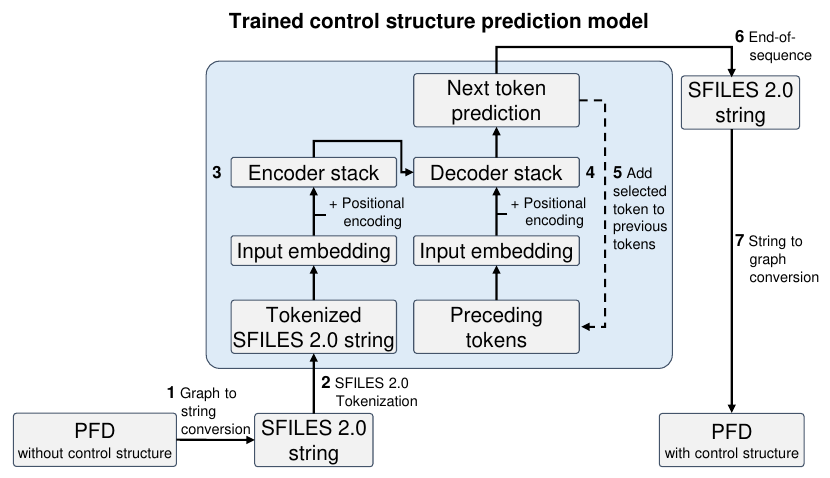}
         \caption{}
         \label{fig:OverviewModel}
     \end{subfigure}
     \vfill
     \begin{subfigure}[htb]{\textwidth}
         \centering
         \includegraphics[page=8, scale=0.85]{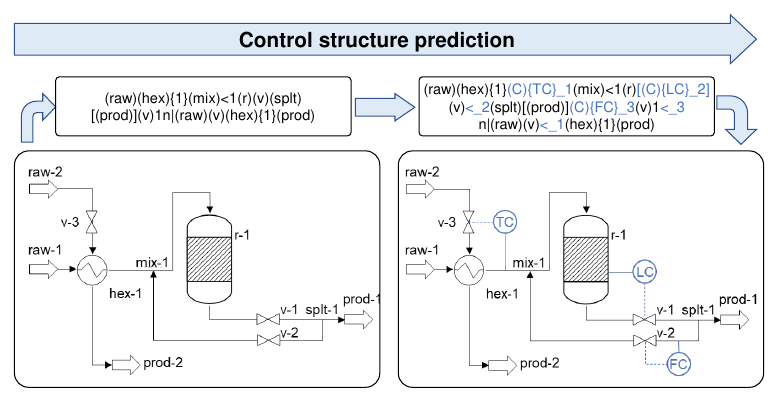}
         \caption{}
         \label{fig:Example_PID_Prediction}
     \end{subfigure}
     \caption{Overview of the control structure prediction with the transformer model. (a)~Conversion of the PFD to SFILES~2.0~(1). Processing of the input SFILES~2.0 with transformer model to predict the control structure~(2-5). Conversion of the output SFILES~2.0 to the PFD including the corresponding control structure~(6-7). (b)~Example control structure prediction adapted from~\cite{Vogel.2022}}
\end{figure}

Figure~\ref{fig:OverviewModel} presents an overview of the control structure prediction model, which is described in the following.
Firstly, the PFD, which is subject to the development of a control structure, is converted to the corresponding SFILES~2.0 string as described in Section~\nameref{subsubsec:SFILES2} (Step~1).
Then, the SFILES~2.0 string is split into chunks of text using the SFILES~2.0 tokenizer as explained in Section~\nameref{subsec:Tokenization} (Step~2).
After converting the tokenized string to an input embedding and adding a positional encoding, the encoder stack computes a numerical embedding of the input string (Step~3).
In Step~4, the decoder stack is initially prompted with a start-of-sequence token.
In combination with the numerical embedding of the input sequence produced by the encoder, the decoder stack predicts the next token of the output sequence.
The predicted token is then added to the preceding tokens of the output sequence and the decoder is again prompted to predict the next token (Step~5).
This auto-regressive prediction of tokens is continued until an end-of-sequence token terminates the prediction process (Step~6).
Lastly, the resulting SFILES~2.0 string is converted to its corresponding graph representation, the PFD with control structure.
Eventually, this procedure could be implemented in CAD software packages to automatically generate the control structure of a drawn PFD as depicted in Figure~\ref{fig:Example_PID_Prediction}.
\newpage
\newpage

\subsection{Tokenization}
\label{subsec:Tokenization}

Tokenizers are generally used in NLP to split text sequences into pieces that can be processed by the language model.
The aim is to compress as many words of a language as possible into a fixed vocabulary while preserving the meaning of the words.
Using the vocabulary, tokenizers convert the input sequence into a numerical vector, which can be processed by the NLP model.
Different tokenization algorithms have been developed according to different languages and intended use cases.
The most commonly used tokenization algorithms comprise word- and subword-based tokenizers, which split the text into words or parts of words and automatically build their vocabulary.
Examples of popular subword-based tokenizers include Byte-pair encoding~(BPE)~\cite{Gage.1994} and SentencePiece~\cite{Kudo.2018}. 

We perform the tokenization of the SFILES~2.0 using a custom tokenizer to preserve the inherent structure of the SFILES~2.0 notation, which is significantly different from natural language.
Inspired by the SMILES tokenizer of Schwaller~et~al.~\cite{Schwaller.2018}, we propose a SFILES~2.0 tokenizer, which, for instance, identifies unit operations (e.g., (hex)), stream tags (e.g., {tout}), letter codes (e.g., {LC}), material recycle (e.g., <\#, \#) and signal connections (e.g., <\_\#, \_\#).
The SFILES~2.0 string is split into pieces using the following regular expression, which is used to search and match certain patterns in the SFILES~2.0:

\begin{center}
    \texttt{(\textbackslash(.+?\textbackslash)|\textbackslash\{.+?\textbackslash\}|[<\%\_]+\textbackslash d+|\textbackslash]|\textbackslash[|\textbackslash<\textbackslash\&\textbackslash||(?<!<)\&\textbackslash||n\textbackslash||(?<!\&)(?<!n)\textbackslash||\&(?!\textbackslash|)|\textbackslash d)}.
\end{center}

For example, tokenizing the following SFILES~2.0 string

\begin{center}
    \texttt{(raw)(hex)\{1\}(C)\{TC\}\_1(mix)<1(r)[(C)\{LC\}\_2](v)<\_2(splt)[(prod)]\\(C)\{FC\}\_3(v)1<\_3n|(raw)(v)<\_1(hex)\{1\}(prod)}
\end{center}

results in

\begin{center}
   \texttt{(raw), (hex), \{1\}, (C), \{TC\}, \_1, (mix), <1, (r), [, (C), \{LC\}, \_2, ], (v), <\_2, (splt), [, (prod), ], (C), \{FC\}, \_3, (v), 1, <\_3, n|, (raw), (v), <\_1, (hex), \{1\}, (prod)}.
\end{center}

\subsection{T5-Model for control structure prediction}
\label{subsec:T5}

There exist multiple, publicly available sequence-to-sequence models, such as OpenNMT~\cite{klein-etal-2017-opennmt} or the T5 model~\cite{Raffel.2019}.
We use the T5 transformer model~\cite{Raffel.2019}, a state-of-the-art model easily accessible through \href{https://huggingface.co}{Hugging Face}, casting the control structure prediction as a translation task.
Therefore, the employed model is a sequence-to-sequence model with an encoder-decoder structure as explained in Section~\nameref{subsec:Seq2Seq}.
The T5 model is in large parts equivalent to the original transformer architecture proposed by Vaswani~et~al.~\cite{Vaswani.2017}.
Modifications include removing the layer bias norm, placing layer normalization outside the residual connections, and applying a different positional encoding~\cite{Raffel.2019}.
Since the SFILES~2.0 vocabulary is limited to a few hundred entries, we utilize the T5-small version with originally about 60~million parameters.
The T5-small model has an embedding size of 512, utilizes an 8-headed attention mechanism, and consists of six encoder and decoder layers each.
Preliminary tests on a generated SFILES~2.0 dataset with around 10,000 samples indicate, that an even smaller architecture may be sufficient and advantageous.
For this reason, we further decrease the model size of the T5-small model by reducing the embedding size to 128 and the number of encoder and decoder layers each to two.
In summary, our model comprises roughly 7.9~million trainable parameters.
Compared to other state-of-the-art language models, our model size is significantly smaller, but our vocabulary sizes are clearly smaller, too. We performed a hyperparameter optimization including the model size and several other model and training parameters based on a grid-search hyperparameter tuning run~(cf. Supplementary Information Table~S1). The results of the hyperparameter tuning run suggest that smaller models lead to better results. Thus, the consideration of even simpler model architectures, such as simple RNNs, in addition to extensive hyperparameter tuning runs could be promising for future work.
During model training, early-stopping is utilized to prevent overfitting and unnecessary long training runs.
Evaluation of the model is performed by generating predictions with beam search as decoding strategy as described in Section~\nameref{subsec:Seq2Seq}.
The beam width is set to five and those five, most probable predictions are returned to the user as recommendations for possible control structures of the provided PFD.
The implementation of a constrained beam search would be possible to prevent the model from predicting unit operations, which are not present in the PFD.
However, such an implementation is not applied in the following experiments.

\section{Results and discussion}
\label{sec:Results}

This section summarizes the training procedure for pre-training and fine-tuning the control structure prediction model.
Thereafter, the model is evaluated based on the top-\textit{k} accuracy metric.

\subsection{Model training}
\label{subsec:ModelTraining}

We perform model pre-training with different generated training set sizes as denoted in Table~\ref{tab:top_k_pretraining}.
Additionally, an independent validation and test set is generated with 1000 samples each.
During pre-training, we use a learning rate of $3\cdot10^{-4}$ and a batch size of 32.
Model evaluation is performed depending on the dataset size every 500 steps for the training dataset containing 10,000 and 100,000 samples, every 25 steps for the dataset with 1,000 samples, and every 5 steps for the dataset with 100 samples.
Early stopping is applied with patience of 10 steps to prevent overfitting.

Subsequently, we fine-tune the pre-trained model on real PFDs with control structures splitting the dataset into a train~(80\%), validation~(10\%), and test~(10\%) set.
Model fine-tuning is performed with a reduced learning rate of $0.5\cdot10^{-4}$ and a batch size of 2.
We evaluate the model every 20 steps and apply early stopping with patience of 40 steps.

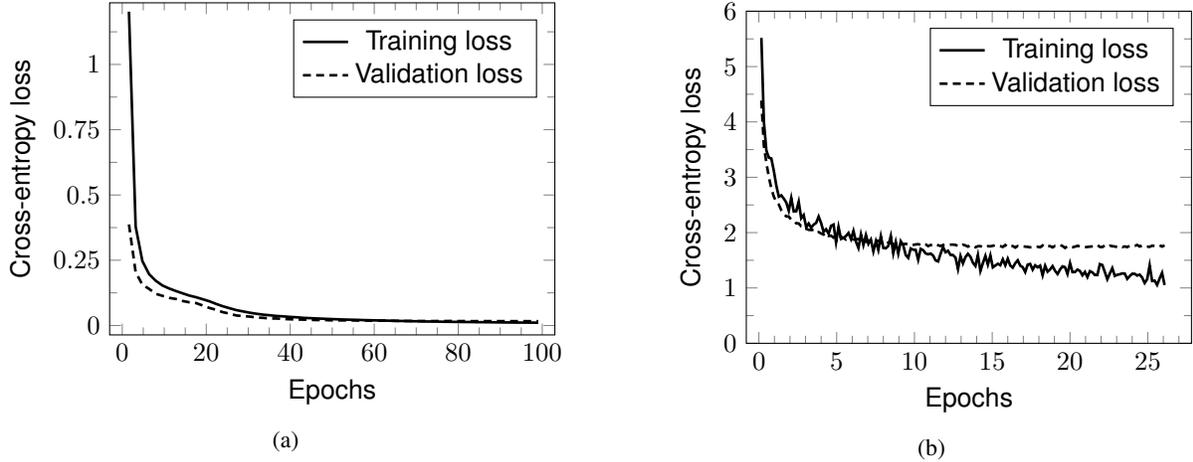
\begin{figure}[p!]
    \centering
    \begin{subfigure}[htb]{0.55\textwidth}
        \centering
        \begin{tikzpicture}[trim axis left]
            \begin{axis}[
        	    xlabel={Epochs},
        		xmin=0, xmax=100,
        		minor x tick num=3,
        		enlarge x limits=0.03,
        		enlarge y limits=0.03,
        		ylabel={Cross-entropy loss},
        		ylabel near ticks,
        		width=7.5cm, height=6cm,
        		ymin=0, ymax=1.2,
        		ytick={0, 0.25,  0.5, 0.75, 1, 1.25},
        		minor y tick num=1,
        		legend style={at={(0.96,0.95)},anchor=north east}
        		]
        		
        		\addplot [black, line width=1pt] table[x=epochs, y=loss_train] {figures_acceptedversion/history_pretraining.txt};
        		\addlegendentry{Training loss};
        		\addplot [black, line width=1pt, densely dashed] table[x=epochs, y=loss_val] {figures_acceptedversion/history_pretraining.txt};
        		\addlegendentry{Validation loss};
        		
            \end{axis}
        \end{tikzpicture}
        \caption{}
        \label{fig:loss_pretraining}
    \end{subfigure}
    \hfill
    \begin{subfigure}[htb]{0.55\textwidth}
        \centering
        \begin{tikzpicture}[trim axis left]
            \begin{axis}[
        	    xlabel={Epochs},
        		xmin=0, xmax=27,
        		minor x tick num=4,
        		enlarge x limits=0.03,
        		ylabel={Cross-entropy loss},
        		ylabel near ticks,
        		width=7.5cm, height=6cm,
        		ymin=0, ymax=6,
        		ytick={0, 1, 2, 3, 4, 5, 6},
        		minor y tick num=1,
        		legend style={at={(0.96,0.95)},anchor=north east}
        		]
        		
        		\addplot [black, line width=1pt] table[x=epochs, y=loss_train] {figures_acceptedversion/history_finetuning.txt};
        		\addlegendentry{Training loss};
        		\addplot [black, line width=1pt, densely dashed] table[x=epochs, y=loss_val] {figures_acceptedversion/history_finetuning.txt};
        		\addlegendentry{Validation loss};
        		
            \end{axis}
        \end{tikzpicture}
        \caption{}
        \label{fig:loss_finetuning}
    \end{subfigure}
    \caption{Training and validation loss curve during (a)~pre-training with 10,000 training samples and (b)~fine-tuning}
    \label{fig:loss}
\end{figure}

Figure~\ref{fig:loss_pretraining} illustrates exemplary the training and validation loss curves during model pre-training with a dataset size of 10,000 generated PFDs with decentralized control structures.
The first few epochs exhibit a steep decrease of both training and validation loss, whereupon the losses in the subsequent epochs asymptotically approach a constant value.
The gap between training and validation loss is small, indicating a small generalization error, which is likely due to the limited variance in the generated dataset.
Additionally, the samples of training and validation set are drawn from the same probability distribution and thus forming a representative validation set.
The early stopping callback detects no increase in the validation loss during model training. The small difference between training and validation loss is an indication that overfitting is not observed.

Figure~\ref{fig:loss_finetuning} depicts the training and validation loss curves during model fine-tuning.
Compared to Figure~\ref{fig:loss_pretraining} a larger gap between training and validation loss curve and generally higher fluctuations are observed.
This behavior is most likely related due to the training on real PFDs with control structures, which generally exhibit a higher complexity than the generated examples.
As indicated in Table~\ref{tab:Dataset_Properties}, the real data shows higher variations in the number of nodes and due to additional other unit operations and letter codes in the control structures, resulting in an extended vocabulary size.
Along with the small dataset size, the validation set is likely not representative.
The early stopping callback detects an increase in the validation loss at around epoch 27 and thus terminates the model training run. The difference between training and validation loss is due to higher variations in the dataset within an expected range but still in the same order of magnitude.
The experiments with different dataset sizes during pre-training resulted in qualitatively similar loss curves during pre-training and fine-tuning.

\subsection{Model evaluation}
\label{subsec:Results}

The model performance after pre-training on different generated dataset sizes is evaluated based on the top-\textit{k} accuracy.
Therefore, the top-5 predictions are determined with beam search decoding.
A prediction is counted as true, if the target PFD with control structure is present in the top-\textit{k} predictions of the model. 
The results, presented in Table~\ref{tab:top_k_pretraining}, show that increasing the dataset size significantly improves the model performance.
It is evident that a dataset size of 100 or 1,000 samples is not sufficient for pre-training the control structure prediction model.
With 10,000 generated process diagrams, we already reach a top-5 accuracy of roughly 75\% on the test set consisting of generated data during pre-training.
The top-5 accuracy can be increased up to 89.2\% on the test set, consisting of generated data, when pre-training with 100,000 generated samples. 
Therefore we conclude, that the control structure prediction model learns the grammatical structure of SFILES~2.0, which are based on generated PFDs, and correctly gives recommendations for the control structure of unknown PFDs, corresponding to the generated dataset, through learning the patterns present in the training data.
The transformer model learns conditional probabilities and essentially exploits common patterns in the data.
It is highlighted that this generated dataset is only based on topological information of the PFDs and no information such as process dynamics, operating points, or equipment sizing is provided therein.
Since this generated dataset is created under different assumptions than real PFDs with control structure, these results are not directly transferable to real conditions.
Nevertheless, the results indicate successful learning of our proof-of-concept on synthetic data.
In addition, the results indicate, that SFILES~2.0 data augmentation has positive effects on the model performance.
Especially on the dataset with 10,000 samples, a significant increase in the top-1 accuracy is observed after augmentation.

Since valves are often omitted in PFDs, an additional pre-training run is performed with 10,000 training samples, where the entire control structure and all valves are removed from the input dataset.
Thus, the model learns to predict not only the control structures but also the valves. 
The results, denoted in Table~\ref{tab:top_k_pretraining}, indicate that it is significantly more difficult for the model to predict correct control structures.
This causes the top-1 accuracy to decrease from 37.7\% (10,000 input samples with valves) to 17.8\% (10,000 input samples without valves).
However, this demonstrates that the model is also capable of predicting correct valve positions, which are not necessarily present in the PFDs.

\begin{table}[p!]
	\caption{Top-\textit{k} accuracy of the pre-trained model on the generated test set}
	\centering
	\begin{threeparttable}
    	\begin{tabular}{lccccc}
    		\toprule
    		Samples in\\ training data & top-1~(\%) & top-2~(\%) & top-3~(\%) & top-4~(\%) & top-5~(\%)\\ 
    		\midrule
    		100 & 0.0 & 0.0 & 0.0 & 0.0 & 0.0\\
    		199\textsuperscript{*} & 0.0 & 0.0 & 0.0 & 0.0 & 0.0\\
    		1,000 & 0.3 & 0.6 & 0.6 & 0.8 & 0.9\\
    		1,958\textsuperscript{*} & 0.3 & 0.3 & 0.4 & 0.5 & 0.6\\
    		10,000 & 37.7 & 56.3 & 65.8 & 72.0 & 74.8\\
    		10,000\textsuperscript{**} & 17.8 & 30.2 & 38.9 & 44.2 & 48.0 \\
    		19,573\textsuperscript{*} & 41.4 & 61.1 & 67.7 & 73.7 & 76.1\\
    		100,000 & 48.6 & \textbf{71.3} & \textbf{81.4} & \textbf{86.7} & \textbf{89.2}\\
    		195,467\textsuperscript{*} & \textbf{49.7} & 70.6 & 80.9 & 85.7 & 87.5\\
    		\bottomrule
    	\end{tabular}
    	
    	\begin{tablenotes}
            \small
            \item \begin{flushleft}
                \textsuperscript{*}augmented dataset
            \end{flushleft}
            \item \begin{flushleft}
                \textsuperscript{**}removed valves in input training data
            \end{flushleft}
        \end{tablenotes}
    \end{threeparttable}
    \label{tab:top_k_pretraining}
\end{table}

In a first experiment, we trained the control structure prediction model directly on 250 real PFDs with control structures.
This approach did not yield useful results, as apparently the size of the dataset of 250 real P\&IDs is not sufficient to train a transformer-based NLP model.
In a second experiment, we applied a transfer learning method.
We fine-tuned the control structure prediction model with real PFDs with control structures using checkpoints obtained from pre-training with generated data.
Still, the results after fine-tuning revealed a top-5 accuracy of 0\% on the test set of real PFDs with control structures.
Therefore, based on the top-\textit{k} metric, we cannot demonstrate the utility of the model on real data.
While the trained model is not yet applicable to industrial applications, the result is consistent with the results from pre-training.
In particular, the pre-training on a small number of generated PFDs with control structures indicates, that 100 or even 1,000 training samples are not sufficient for reasonable results (cf. Table~\ref{tab:top_k_pretraining}).
The pre-training results highlight, that a sufficiently large number (here, 10,000) of training PFDs with control structures is necessary to enable the model learning patterns in the provided data.
Depending on the complexity of real PFDs with control structures the amount of required training samples may be significantly larger due to the increasing complexity of the task.

Error sources and difficulties for the control structure prediction model are not only due to the small real dataset size but also to the dataset composition.
The PFDs with control structures are derived from scientific literature and publicly available sources representing laboratory setups, but also chemical plants or fictive examples and may contain errors, incomplete control structures, and wrong, not standardized, letter codes.
In addition, the real dataset, as described in Section~\nameref{subsec:RealData}, contains very heterogeneous and generally more complex PFDs with control structures.
In combination with the small size of the dataset, this leads to errors in the model predictions, including added or missing unit operations, invalid SFILES~2.0, not connected material recycles, or signal connections.
These errors could be partly mitigated by implementing a constrained beam search algorithm, which sets the probabilities of unit operations not present in the input sequence to zero and forces the model to add only the control structure and valves to the output sequence.
Nevertheless, since for every section of a chemical plant exists at least one PFD and P\&ID, we believe that there is enough data available in the proprietary domain to train our control structure prediction model making no arbitrarily changes in the PFD and predicting correct control schemes.
Here, a major limitation is the availability of real flowsheet data to academia, as most of the PFDs and P\&IDs in the industry are protected confidential.

\subsection{Illustrative example}
\label{sec:Examples}

This section illustrates the model predictions on one representative sample taken from the independent test set.
For this illustrative example, we use the model that has been trained on 10,000 training samples without data augmentation and fine-tuning.
The model is prompted with a PFD (colored black in Figure~\ref{fig:IllustrativeExample}) of the test dataset, as denoted in the following SFILES~2.0 string:

\begin{flushleft}
    \leftskip5mm
    \texttt{(raw)(hex)\{1\}(mix)<\&|(raw)(v)\&|(v)(hex)\{2\}(rect)<1<2[\{tout\}\\(cond)(sep)[(v)(prod)](splt)[(v)(prod)](v)1]\{bout\}(splt)[(v)\\
    (prod)](hex)\{3\}2n|(raw)(hex)\{1\}(v)(prod)n|(raw)(v)(hex)\{2\}\\
    (prod)n|(raw)(v)(hex)\{3\}(prod)}.
\end{flushleft}

The model predicts with beam search decoding the following five, syntactically correct SFILES~2.0.
These SFILES~2.0 contain the input PFD colored in black and the predicted, five most-likely control structures illustrated in blue:

\begin{enumerate}
    \item \texttt{(raw)(hex)\{1\}\tb{(C)\{TC\}\_1(C)\{FT\}\_2}(mix)<\&|(raw)\tb{(C)\{FFC\}\_3<\_2}(v)\&\\
    \tb{<\_3}|\tb{(C)\{FC\}\_4}(v)\tb{<\_4}(hex)\{2\}\tb{(C)\{TC\}\_5}(rect)<1<2\tb{[(C)\{PC\}\_6][(C)\\
    \{LC\}\_7]}[\{tout\}(cond)(sep)\tb{[(C)\{LC\}\_8]}[(v)\tb{<\_6}(prod)](splt)[(v)\\
    \tb{<\_8}(prod)]\tb{(C)\{FC\}\_9}(v)1\tb{<\_9}]\{bout\}(splt)[\tb{(C)\{FC\}\_10}(v)\tb{<\_10}\\
    (prod)](hex)\{3\}2n|(raw)(hex)\{1\}(v)\tb{<\_1}(prod)n|(raw)(v)\tb{<\_5}\\(hex)\{2\}(prod)n|(raw)(v)\tb{<\_7}(hex)\{3\}(prod)}
    \item \texttt{(raw)(hex)\{1\}\tb{(C)\{TC\}\_1(C)\{FT\}\_2}(mix)<\&|(raw)\tb{(C)\{FFC\}\_3<\_2}(v)\&\\
    \tb{<\_3}|\tb{(C)\{FC\}\_4}(v)\tb{<\_4}(hex)\{2\}\tb{(C)\{TC\}\_5}(rect)<1<2\tb{[(C)\{PC\}\_6][(C)\\
    \{LC\}\_7][(C)\{TC\}\_8]}[\{tout\}(cond)(sep)\tb{[(C)\{LC\}\_9]}[(v)\tb{<\_6}(prod)]\\
    (splt)[(v)\tb{<\_9}(prod)]\tb{(C)\{FC\}\_10}(v)1\tb{<\_10}]\{bout\}(splt)[\tb{(C)\{FC\}\\
    \_11<\_8}(v)\tb{<\_11}(prod)](hex)\{3\}2n|(raw)(hex)\{1\}(v)<\tb{\_1}(prod)\\
    n|(raw)(v)\tb{<\_5}(hex)\{2\}(prod)n|(raw)(v)\tb{<\_7}(hex)\{3\}(prod)}
    \item \texttt{(raw)(hex)\{1\}\tb{(C)\{TC\}\_1(C)\{FT\}\_2}(mix)<\&|(raw)\tb{(C)\{FFC\}\_3<\_2}(v)\&\\
    \tb{<\_3}|\tb{(C)\{FC\}\_4}(v)\tb{<\_4}(hex)\{2\}\tb{(C)\{TC\}\_5}(rect)<1<2\tb{[(C)\{PC\}\_6][(C)\\
    \{LC\}\_7][(C)\{TC\}\_8]}[\{tout\}(cond)(sep)\tb{[(C)\{LC\}\_9]}[(v)<\tb{\_6}(prod)]\\
    (splt)[\tb{(C)\{FC\}\_10<\_8}(v)<\tb{\_10}(prod)](v)1\tb{<\_9}]\{bout\}(splt)[(v)<\tb{\_7}\\
    (prod)](hex)\{3\}2n|(raw)\tb{(C)\{FC\}\_11}(v)\tb{<\_11}(hex)\{3\}(prod)\\
    n|(raw)(hex)\{1\}(v)\tb{<\_1}(prod)n|(raw)(v)\tb{<\_5}(hex)\{2\}(prod)}
    \item \texttt{(raw)\tb{(C)\{FC\}\_1}(v)\tb{<\_1}(mix)<\&|(raw)(hex)\{1\}\tb{(C)\{TC\}\_2(C)\{FT\}}\&\tb{\_3}|\\
    \tb{(C)\{FC\}\_4}(v)<\tb{\_4}(hex)\{2\}\tb{(C)\{TC\}\_5}(rect)<1<2\tb{[(C)\{PC\}\_6][(C)\\
    \{LC\}\_7]}[\{tout\}(cond)(sep)\tb{[(C)\{LC\}\_8]}[(v)\tb{<\_6}(prod)](splt)[(v)\\
    \tb{<\_8}(prod)]\tb{(C)\{FC\}\_9}(v)1\tb{<\_9}]\{bout\}(splt)[\tb{(C)\{FC\}\_10}(v)\tb{<\_10}\\
    (prod)](hex)\{3\}2n|(raw)(hex)\{1\}(v)\tb{<\_2}(prod)\\
    n|(raw)(v)\tb{<\_5}(hex)\{2\}(prod)n|(raw)(v)\tb{<\_7}(hex)\{3\}(prod)}
    \item \texttt{(raw)(hex)\{1\}\tb{(C)\{TC\}\_1(C)\{FT\}\_2}(mix)<\&|(raw)\tb{(C)\{FFC\}\_3<\_2}(v)\&\\
    \tb{<\_}3|\tb{(C)\{FC\}\_4}(v)\tb{<\_4}(hex)\{2\}\tb{(C)\{TC\}\_5}(rect)<1<2\tb{[(C)\{PC\}\_6][(C)\\
    \{LC\}\_7]}[\{tout\}(cond)(sep)\tb{[(C)\{LC\}\_8]}[(v)\tb{<\_6}(prod)]
    (splt)[\tb{(C)\\
    \{FC\}\_9}(v)\tb{<\_9}(prod)](v)1\tb{<\_8]}\{bout\}(splt)[(v)\tb{<\_7}(prod)](hex)\{3\}\\
    2n|(raw)\tb{(C)\{FC\}\_10}(v)\tb{<\_10}(hex)\{3\}(prod)\\
    n|(raw)(hex)\{1\}(v)\tb{<\_1}(prod)n|(raw)(v)\tb{<\_5}(hex)\{2\}(prod)}
\end{enumerate}

Figure~\ref{fig:IllustrativeExample} illustrates the five model predictions.
The PFD, colored black in Figure~\ref{fig:IllustrativeExample}, contains two feed pre-heater, a mixing point of two material streams, and a distillation column.
The model predicts a temperature-dependent control of the utility stream for both feed pre-heater.
Mixing of the two raw material streams is, according to the model, most likely done with a flow ratio control.
Furthermore, the model provides correct predictions of four different distillation column control schemes, which are included in the seven column control structures used to generate the data.
The first prediction~(Figure~\ref{fig:Prediction_1}) corresponds to the ground truth for the corresponding PFD fed to the model as input.

Apart from the correct predictions, Figure~\ref{fig:IllustrativeExample} illustrates limitations and errors of the control structure prediction model.
In Figure~\ref{fig:Prediction_4}, the model inserts a flow transmitter and fails to predict a corresponding signal connection.
In addition, the mixing of the material flows upstream of the distillation column could be problematic from a control perspective, as flow control is proposed here before and after mixing.
This problem arises from model training with generated data, which is synthesized by adding small control patterns to a final PFD with control structure.
The addition of the utilized control patterns may not result in a meaningful, correct control architecture, and furthermore, no long-range dependencies but only decentralized control structures are considered in the data generation procedure.
In addition, correct predictions of the control structure of a distillation column setup depend not only on the topological structure, but also depend on additional information such as present components, material flows, or operating conditions.
This means that control structure predictions considering only the topological structure of the process may appear correct, but are wrong when considering in detail light- and heavy-boiling components, azeotropic mixtures, or quality measures.

\begin{figure}[p!]
    \centering
    \begin{subfigure}[]{\textwidth}
        \centering
        \includegraphics[scale=0.7,page=1]{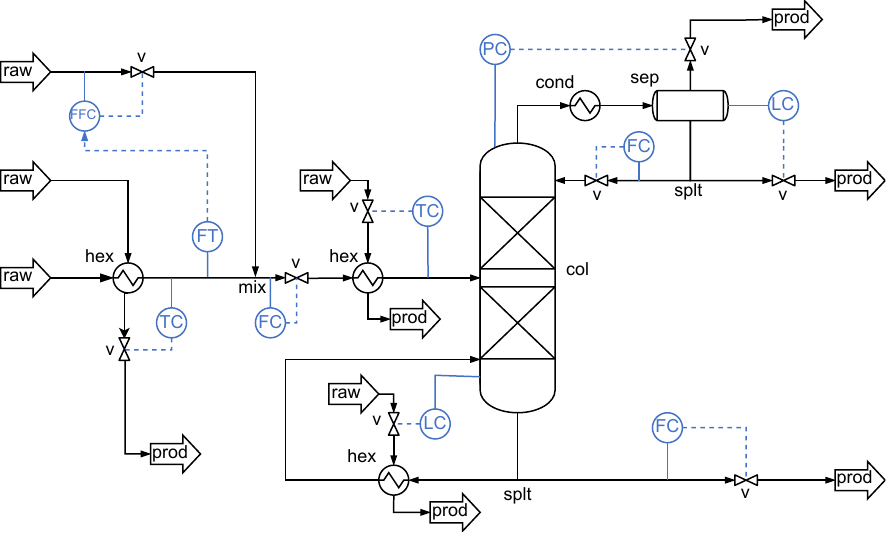}
        \caption{Prediction~1}
        \label{fig:Prediction_1}
    \end{subfigure}
    \begin{subfigure}[]{\textwidth}
        \centering
        \includegraphics[scale=0.7,page=2]{figures_acceptedversion/IllustrativeExample.pdf}
        \caption{Prediction~2}
        \label{fig:Prediction_2}
    \end{subfigure}
\end{figure}
\begin{figure}[p!]\ContinuedFloat
    \begin{subfigure}[]{\textwidth}
        \centering
        \includegraphics[scale=0.7,page=3]{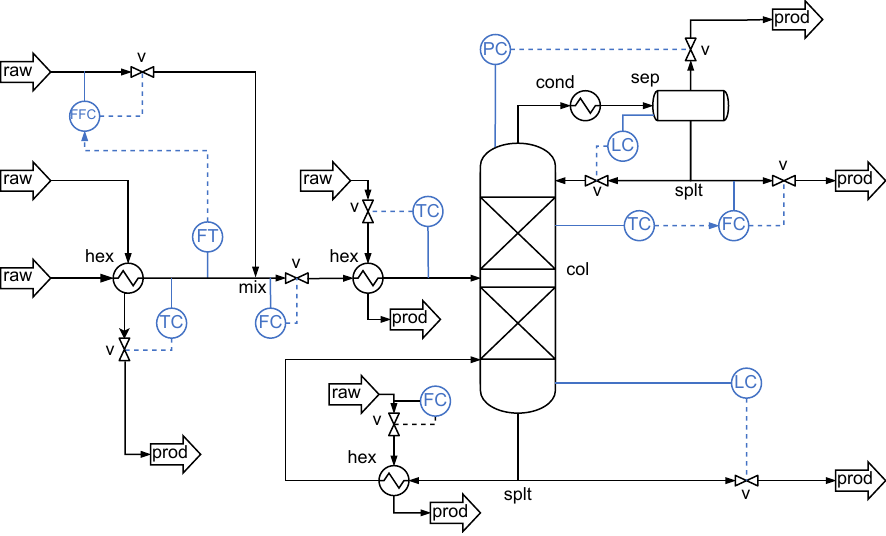}
        \caption{Prediction~3}
        \label{fig:Prediction_3}
    \end{subfigure}
    \begin{subfigure}[]{\textwidth}
        \centering
        \includegraphics[scale=0.7,page=4]{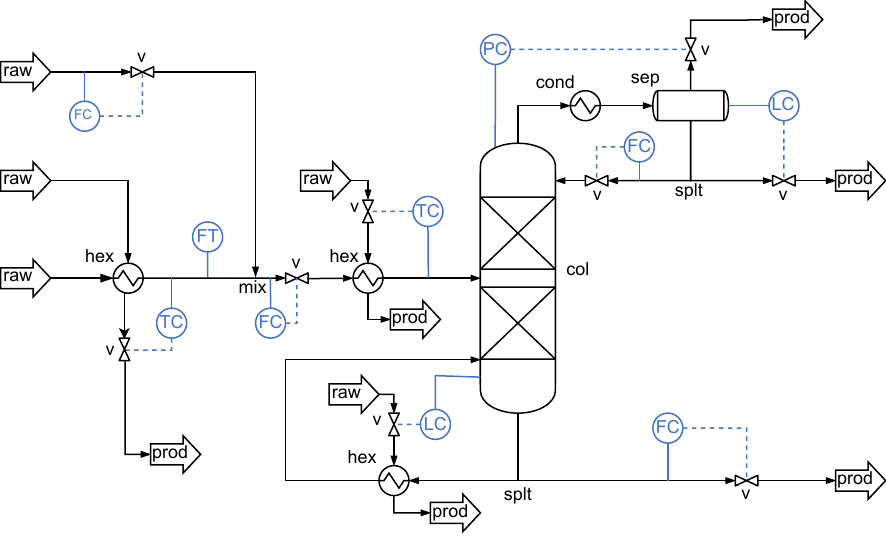}
        \caption{Prediction~4}
        \label{fig:Prediction_4}
    \end{subfigure}
    \begin{subfigure}[]{\textwidth}
        \centering
        \includegraphics[scale=0.7,page=5]{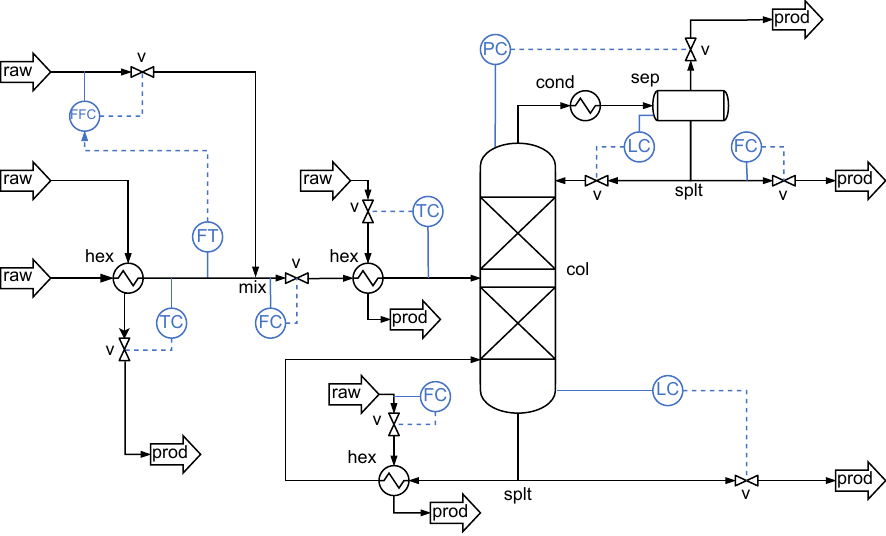}
        \caption{Prediction~5}
        \label{fig:Prediction_5}
    \end{subfigure}
    \caption{Control structure predictions~(a)-(e) (in blue) of the model prompted with the PFD (colored black) as input}
    \label{fig:IllustrativeExample}
\end{figure}
\begin{figure}[p!]
	\centering
	\includegraphics[scale=0.7,page=7]{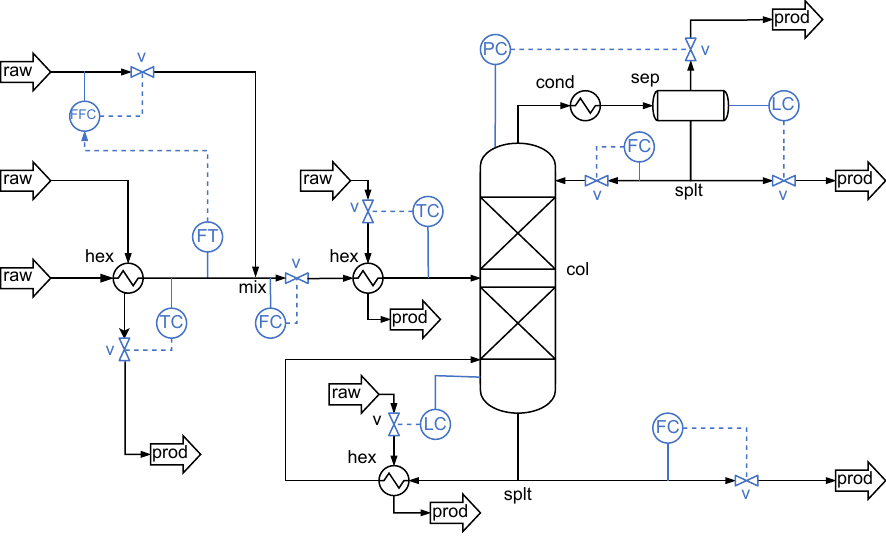}
	\caption{Control structure prediction (in blue) of the model prompted with the PFD (colored black) as input}
	\label{fig:PID_Generation_wo_valves}
\end{figure}

\newpage
Since PFDs do not necessarily contain valves, we trained a model with 10,000 training samples, removing not only the entire control structure but also each valve in the input data.
Given the PFD (colored black in Figure~\ref{fig:PID_Generation_wo_valves}) and excluding any valve in the model input, the third prediction of the model, as illustrated in Figure~\ref{fig:PID_Generation_wo_valves}, represents the ground truth.
This shows that our model has also the potential to learn the positioning of valves in combination with the prediction of the control structure.

\subsection{Current limitations and future directions}
\label{subsec:Current limitations}

Overall, our results show great potential for automatically predicting decentralized control structures for PFDs. 
However, the results also demonstrate several current limitations that need to be overcome for industry applications.
A main issue related to the proposed model is the consideration of only topological information of the PFD for the prediction of an appropriate control structure.
As already discussed in Section~\nameref{sec:Examples} specifically for the distillation setup, detailed information about the design characteristics, such as equipment sizing, temperature control (isothermal, adiabatic, non-isothermal), batch or, continuous operation, of every unit operation is necessary to develop a safe and reliable control structure.
In addition, the SFILES~2.0 currently misses information about streams, present components, and operating points.
Furthermore, process dynamics and operating objectives of a chemical plant need to be considered during the design of the control structure.
This missing information severely limits the model performance and could lead to wrong predictions of the control structure.

To overcome the current limitations and challenges, we have identified four main development directions that need to be addressed in order to effectively apply the proposed model to real-world PFDs for control structure prediction in the future.

\begin{enumerate}
    \item \textbf{Data availability:} To train NLP models effectively, a high quantity of training data needs to be accessible for the model. 
    Especially for the prediction of control structures, the collaboration of industry and academia is essential to provide high-quality P\&IDs for model training.
    Ideally, such training data should also be checked/curated thoughtfully.
    As an alternative, we also explore the automatic mining of flowsheets from literature and patents~\cite{SchulzeBalhorn.2022} and their automatic digitization~\cite{Theisen.2022}.

    \item \textbf{Inclusion of additional information:} To encode the required information for a successful control structure design, such as operating points, stream information, or present components, the SFILES~2.0 notation needs to be extended. Another possibility is a switch from the language-based to a graph-based model architecture, where the required information about the chemical process is encoded in a graph instead of a string.
    In our previous work, we showed already that flowsheets and flow information can be represented as graphs~\cite{Stops.2022}. In addition, we already leveraged graph neural networks to learn from these flowsheet graphs~\cite{Stops.2022}. These technologies are also promising in the context of the prediction of decentralized control structures.

    \item \textbf{Hybrid AI solutions:}
    Chemical engineers have developed fundamental principles of modeling and control. Integrating these principles into the future AI algorithm has a great potential to improve extrapolability, increase safety and explainability, and reduce data requirements~\cite{Schweidtmann.2021,Venkatasubramanian.2019}. Thus, this integration is an important future research direction. 
    
    \item \textbf{Validity checks:} Control is safety-critical. Thus, future work requires validity checks for the training data as well as the model predictions. These can be guided by physical knowledge and rules from the process engineering domain~\cite{Venkatasubramanian.2019}, AI approaches~\cite{Oeing.2022}, and hybrid AI approaches~\cite{Schweidtmann.2021}.
\end{enumerate}

Consequently, addressing the mentioned points in future research will make the model applicable to industrial applications.
We envision an integration of our model in CAD software to assist engineers in developing PFDs with control structures.

\FloatBarrier

\section{Conclusion}
\label{sec:Conclusion}

Predicting the control structure of PFDs with machine learning models is a promising strategy to accelerate the development of chemical processes.
We propose a novel method of casting the prediction task as a translation task and leveraging the transformer architecture from the field of NLP.
To apply NLP techniques, we represent the graph-based process diagrams in the text-based SFILES~2.0 notation.
We successfully trained a fully data-driven sequence-to-sequence model to predict the decentralized control structure of generated chemical processes without relying on handcrafted rules.
Experiments on 312 real PFDs with control structures indicate that for reasonable results larger datasets are necessary.

Future work should focus on the acquisition of a larger dataset of real PFDs with control structures, which can be used to fine-tune our model and leverage the possible advantages of transfer learning.
Additionally, the context of the chemical process, such as operating conditions, basic control structures already present in the PFD, or stream information, may be included in advanced models to refine the prediction of the control structure.
Furthermore, training the model on specific classes of plants such as petrochemical or utility systems could be promising since this would decrease the complexity of the prediction task.
Besides the prediction of the control structure, extensions of the control structure prediction model could include e.g., pipe classes or valve types, to finally enable the automatic prediction of complete P\&IDs.  
Moreover, validity checks may be included to further increase the accuracy of the model predictions.
Ultimately, our model should not be seen as an alternative to the control engineer or existing rule-based systems.
According to the UNESCO~\cite{Unesco.2021} AI may assist humans in decision-making for efficiency reasons, but an AI algorithm not replaces human responsibility in safety-critical decisions.
Therefore, we envision a combination of the algorithm with other process development methods to assist the engineer with recommendations, reduce the number of manual tasks, and generally make process development more efficient.
	
	\newpage
	
	\begin{acknowledgements}
This publication is part of the project “ChemEng KG – The Chemical Engineering Knowledge Graph” with project number 203.001.107 of the research program “Open Science (OS) Fund 2020/2021” which is (partly) financed by the Dutch Research Council (NWO).
We want to thank the anonymous reviewers for their time and effort in evaluating our work. This greatly helped us to clarify and improve our publication.
	\end{acknowledgements}
	
    \bibliographystyle{ama}
    \bibliography{references}   

    \newpage
    \listoffigures
\end{document}